%% file: main.tex
\pdfoutput=1

\documentclass[11pt]{article}

\usepackage{acl}

\usepackage{times}
\usepackage{latexsym}
\usepackage[T1]{fontenc}
\usepackage[utf8]{inputenc}
\usepackage{microtype}
\usepackage{inconsolata}
\usepackage{graphicx}
\usepackage{pifont}
\usepackage[utf8]{inputenc} 
\usepackage[T1]{fontenc}    
\usepackage{hyperref}       
\usepackage{times}
\usepackage{url}            
\usepackage{booktabs}       
\usepackage{amsfonts}       
\usepackage{nicefrac}       
\usepackage{microtype}      
\usepackage{xcolor}         

\usepackage{float}      
\usepackage{graphicx}
\usepackage{tabularx}
\usepackage{wrapfig}
\usepackage{enumitem}
\usepackage{amsmath}
\usepackage{bm}
\usepackage{bbold}
\usepackage{caption}
\usepackage{multirow}
\usepackage{subcaption}
\usepackage{colortbl}
\usepackage{algorithm}
\usepackage{lipsum}
\usepackage{wrapfig}
\usepackage{amsfonts}
\usepackage{amssymb}
\usepackage{mathtools}
\usepackage{amsthm}
\usepackage{algorithmic}
\usepackage{multirow} 
\usepackage{authblk}

\title{Guiding Medical Vision-Language Models with Explicit Visual Prompts: Framework Design and Comprehensive Exploration of Prompt Variations}

\author{
  Kangyu Zhu\textsuperscript{*}$^{1,4}$, 
  Ziyuan Qin\textsuperscript{*}$^{5}$, 
  Huahui Yi$^{1}$, 
  Zekun Jiang$^{1}$, 
  Qicheng Lao$^{3}$, \\
  Shaoting Zhang$^{6}$, 
  Kang Li$^{1,2}$ \\
  \\
  $^{1}$West China Biomedical Big Data Center, West China Hospital, Sichuan University \\
  $^{2}$Sichuan University Pittsburgh Institute \\
  $^{3}$Beijing University of Posts and Telecommunications
  $^{4}$Brown University\\
  $^{5}$Case Western Reserve University
  $^{6}$Shanghai Jiao Tong University
  
}

\begin{document}

\maketitle
\let\thefootnote\relax  
\footnotetext{\textsuperscript{*}Equal Contribution.}
\begin{abstract}
\input{structure/0_abstract}
\end{abstract}

\input{structure/1_introduction}

\input{structure/2_preliminary}
\input{structure/3_method}

\input{structure/4_experiment}

\input{structure/5_limitation}

\input{structure/6_conclusion}



\bibliography{custom}

\clearpage

\input{structure/appendix}

\end{document}

%% file: structure/0_abstract.tex

While mainstream vision-language models (VLMs) have advanced rapidly in understanding image-level information, they still lack the ability to focus on specific areas designated by humans. Rather, they typically rely on large volumes of high-quality image-text paired data to learn and generate posterior attention maps.
To address this critical issue, we propose leveraging visual prompts—simple visual markers in various forms—to guide and enhance the formation of region-specific attention.
Thus, we introduce \textbf{MedVP}, a pioneering framework that integrates medical entity extraction, visual prompt generation, and dataset adaptation for visual prompt-guided fine-tuning.
We successfully outperform recent state-of-the-art large models across multiple medical VQA datasets. Extensive experiments and Human evaluation are conducted to analyze the impact of different visual prompt forms and how they contribute to performance improvement. The results demonstrate both the effectiveness and clinical significance of our approach.

%% file: structure/1_introduction.tex
\section{Introduction}

As human beings, we inherently rely on visual cues or prompts to understand complex visual content in greater detail. These visual prompts may take various forms, yet they serve a fundamentally similar purpose: to direct our attention toward critical areas that contain rich information.

\begin{figure}[]
  \centering
  \includegraphics[width=\linewidth]{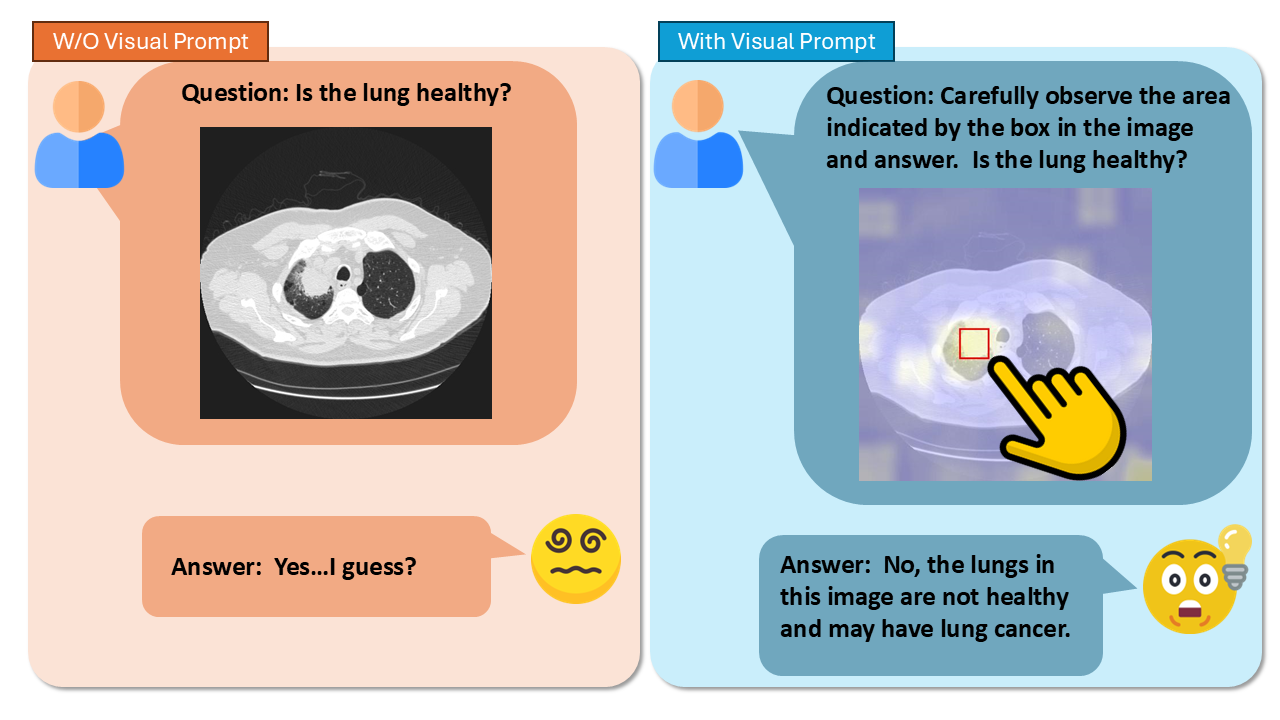}
\caption{The figure illustrates the difference in the reasoning process of the Vision-Language Model (VLM) with and without visual prompts. By emulating how humans approach vision-language tasks, we use an auxiliary visual prompt, akin to pointing at a specific region, to help the model focus more easily on the relevant details and generate accurate responses.}
  \label{fig:fig1}
  \vspace{-1em}
\end{figure}

Current mainstream pre-trained vision-language models often lack mechanisms to direct their attention to specific areas of interest. Instead, their attention distribution is formed during training on specific datasets. However, these pre-trained attention patterns are not universally applicable, particularly in specialized domains such as biomedical images. 
Consequently, a substantial amount of high-quality image-text paired data is required to fine-tune pre-trained VLMs, ensuring their attention distribution aligns with medical image characteristics. This need persists despite the availability of prior knowledge that could guide attention to more meaningful areas. This intuition also aligned with active vision~\citep{karasev} and interactively segmentation~\citep{Tanida_2023} in medical domain.
Therefore, inspired by the human recognition process, we propose leveraging visual prompts—free-form visual markers, such as arrows or circles, to indicate areas of interest—to guide VLMs in attending to specific regions during fine-tuning. This approach facilitates the adaptation of attention distribution and accelerates the formation of posterior attention maps in the medical domain.

To be specific, Vision-Language Models (VLMs) are a class of models designed to process both image and text inputs for cross-modal tasks. Depending on the specific task, the architecture and pre-training strategies of these models can vary significantly. For the purposes of this discussion, we will focus specifically on Large Vision-Language Models (LVLMs), a subset of VLMs that are particularly designed for image-to-language tasks, such as Visual Question Answering (VQA) and Image Captioning.
In this line of research, foundation models such as BLIP-2~\citep{li2023blip} and LLaVA~\citep{liu2024visual} have significantly influenced the design of current mainstream LVLMs. Taking LLaVA as an example, these models are typically composed of an image encoder, a projection layer, and a Large Language Model (LLM)-driven decoder. 

In the majority of Vision-Language Models (VLMs)~\citep{li2023blip, liu2024visual}, pre-training primarily involves generating descriptions of image content, which encourages the model to focus on global visual features rather than specific regions. During this phase, the attended regions in images remain unguided, allowing the attention mechanism to autonomously learn an implicit attention distribution. As previously discussed, conventional self-attention mechanisms in these models often struggle to generate region-specific content. However, certain vision-language tasks, such as Visual Question Answering (VQA) and Referring Expression Comprehension (REC), inherently require the model to attend to specific objects or localized regions. Learning an effective posterior attention distribution for a given textual query, however, typically demands a substantial amount of training data, making it a challenging problem in vision-language learning.

To mitigate this problem, recent research ~\citep{yang2023setofmarkpromptingunleashesextraordinary, rasheed2024glammpixelgroundinglarge, cai2024vip, zhang2024gpt4roiinstructiontuninglarge, kirillov2023segment} has increasingly recognized the value of visual prompts and their application in vision-language tasks. Visual prompts can be categorized as implicit or explicit concerning the visual inputs. In this study, we will focus on explicit visual prompts, which are basically visual markers in various form that can be directly added to the input images.



Inspired by ViP-LLaVA~\citep{cai2024vip}, we decide to add visual markers into the input images to highlight the Region of Interest (ROI) for the visual-prompt guided VLMs to focus on important regions selected by prior knowledge. 
However, current works\citep{zhang2024gpt4roiinstructiontuninglarge, cai2024vip, kirillov2023segment} usually assume these ROIs are provided interactively by human annotators. Given the high cost of acquiring human annotations in medical images, we propose a novel method, \textbf{MedVP} (\textbf{Med}ical \textbf{V}isual \textbf{P}rompting) that will automatically generate visual prompts for the interested region given the context. 
Specifically, our framework first extracts entities or relevant keywords from the queries, as our focus is on medical VQA tasks. Next, we fine-tune an open-set grounding model that generates coordinates for the target area based on the identified medical entities. Finally, these coordinates are used to generate visual prompts in various formats—such as scribbles, bounding boxes, or circles—on the input images. Along with the visual-prompted images, we adapt the text Question-Answer pairs from SLAKE, VQA-RAD, and PMC-VQA datasets to ensure that the language decoder becomes aware of the presence of our visual prompt markers within the images. 
For example, we incorporate text such as "Carefully observe the area in the red box..." to guide the LLM-driven decoder to focus on the visually prompted area, ensuring that the model pays attention to the regions highlighted by the visual prompts.
The visual-prompted images are fine-tuned with Vision-Language Models (VLMs) to develop our model, \textbf{MedVP-LLaVA}—the first medical VLM guided by visual prompts.
We find that our \textbf{MedVP-LLaVA} performs effectively in medical VQA tasks, significantly enhancing performance across multiple datasets.

In conclusion, our MedVP method has at least the following contribution:

\begin{enumerate}
    \item We are the first to introduce explicit visual prompts into medical-specialized Vision-Language Models, and we validate the effectiveness of visual prompts in enhancing performance for medical VQA tasks.
    \item We design a whole framework from extracting keywords from VQA queries to visual grounding given medical entities, generate the visual prompts to the images and adapt three VQA datasets to include the information of our visual prompts.
    \item We then fine-tune the \textbf{MedVP-LLaVA}, a visual-prompt-aware Vision-Language Model (VLM) tailored for the medical imaging domain, and validate its superiority across multiple medical VQA datasets.
    \item We will release the model weights for both \textbf{MedVP-LLaVA} and the medical grounding model, which has been trained on a large dataset. Additionally, we will make our modified VQA datasets publicly available to accelerate research in developing medical VLMs.  
\end{enumerate}

\begin{figure*}[h]
  \centering
  \includegraphics[width=\textwidth]{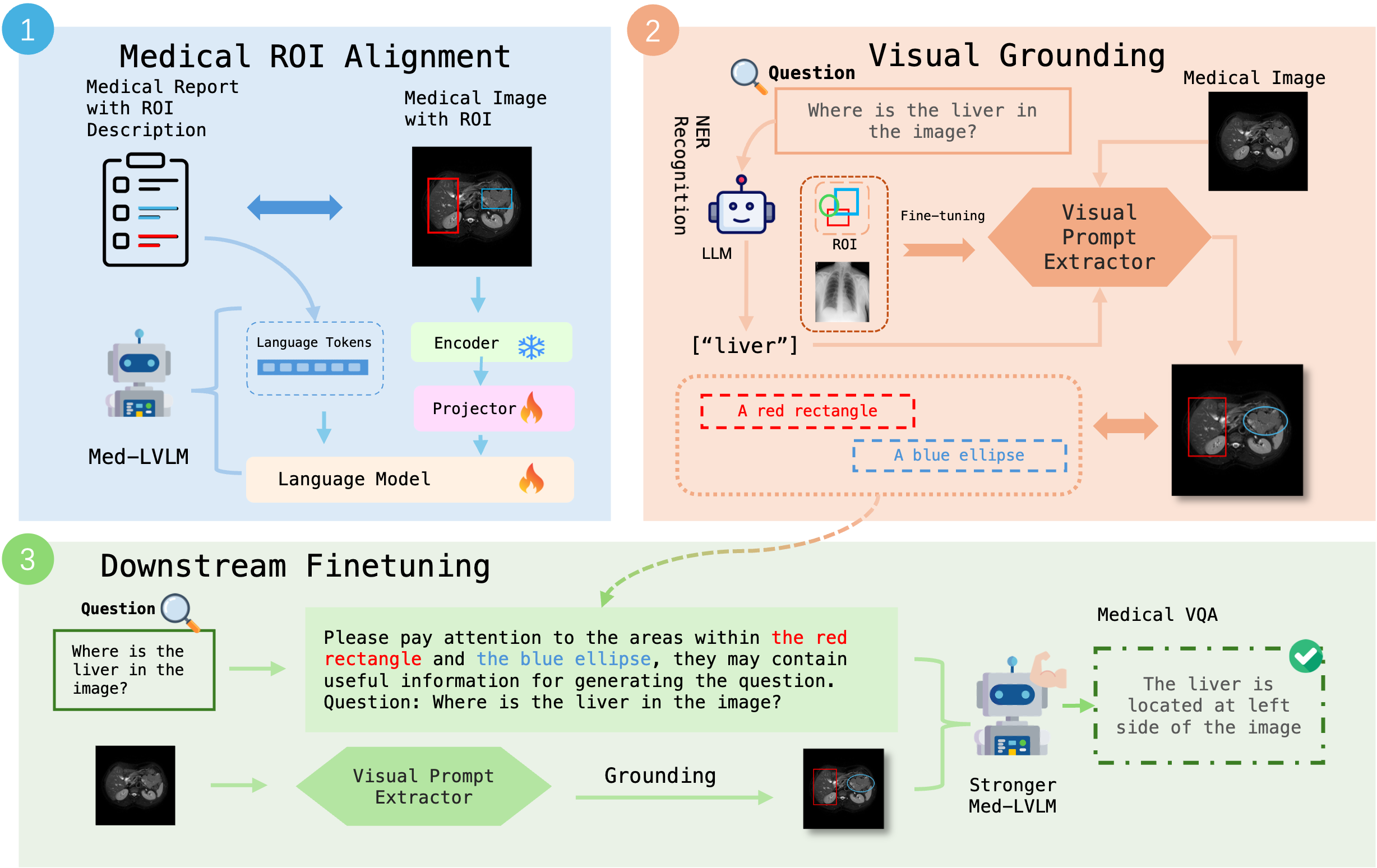}
\caption{The framework of MedVP: an automated, explicit visual prompt-guided approach for medical vision-language models. The framework first aligns the model with region-level medical knowledge. Additionally, it generates explicit visual prompts by leveraging keywords from the question and visual grounding models, integrating these prompts into medical images to enhance performance on medical VQA tasks.}
  \label{fig:framework}
  \vspace{-1em}
\end{figure*}

%% file: structure/2_preliminary.tex
\section{Preliminaries}

\subsection{Medical Vision Language Model}
Medical Large Vision-Language Models (Med-VLMs)~\citep{li2024llava, chen2024huatuogpt, lin2023pmc, xie2024medtrinity, moor2023med, qin2023medicalimageunderstandingpretrained} combine large language models (LLMs) with medical-specific visual modules, allowing them to process medical images alongside clinical queries. When presented with a medical image, the vision encoder extracts visual features, which are then transformed by an adapter module into a format interpretable by the LLM. Utilizing this multimodal input, the model autoregressively predicts the next token in the sequence. A critical capability of these models is Visual Question Answering (VQA), where the model is tasked with answering questions accurately based on the content of the given image.

\subsection{Region-Level Image Understanding}

In most VLMs, the input typically consists of both an image and text, with the model generating responses by integrating the global information from the image and the accompanying text input. However, a common issue arises when the model fails to focus adequately on local regions and fine details, which often leads to incorrect answers. In the domain of natural images, there are three primary approaches to enhancing large models’ attention to local image information, one is incorporating learnable soft tokens into visual inputs for parameter-efficient finetuning~\cite{bahng2022exploring,khattak2023maple}. The second approach is concatenating an image sequence to demonstrate new tasks~\cite{bai2024sequential,bar2022visual}, or using Region of Interest (ROI) features to align language with specific image regions~\cite{guo2024regiongpt}, and the third approach is using grounding regions by overlaying visual markers (such as masks, boxes, or circles) onto the visual inputs~\cite{yao2024cpt,zellers2021merlot}.
In medical imaging, the focus on local regions is particularly crucial, as accurate diagnosis and assessment of abnormalities typically require a combination of both local and global information. However, research on incorporating local region attention in Med-VLMs is still relatively underexplored.

%% file: structure/3_method.tex
\section{Method}
In this section, we outline the proposed method, detailing the training pipeline and objectives for Med-VLMs, including the visual prompt extraction process and fine-tuning downstream medical visual question-answering tasks.
\subsection{Region Level Medical Knowledge Alignment}
Med-VLMs primarily perform alignment training at the image level, as seen in LLaVA-Med~\cite{li2024llava} using visual instruction tuning~\cite{liu2024visual} to align medical context and vision knowledge. However, in medical imaging, diagnostic reasoning often hinges on specific details or localized regions, such as lesions or organs. Failing to adequately focus on these critical areas can lead to incorrect model predictions. Therefore, our training approach ensures that while the model captures global information, it also strengthens the model’s focus and understanding of key regions. Specifically, during instruction tuning for Med-VLMs, we incorporate annotations of local regions in the images, and the corresponding descriptions contain text that explicitly references these annotated areas. 
Therefore, we call these annotations visual prompts for the VLMs since they prompt the models to attend to specific areas.
Each training sample thus consists of three components: the medical image ($\mathbf{I}$), the coordinates of the visual prompts referring to the region of interest within the image ($\mathbf{P}$), and the associated text description ($\mathbf{T}$). The description includes both whole-image-level information and specific details about the region of interest. 
Following the setting in ViP-LLaVA~\citep{cai2024vip}, for the vision part, we merge the image $\mathbf{I}$ and its visual prompts using alpha blending, generating a composite representation that emphasizes the relevant local areas.
\begin{equation}
\hat{\mathbf{I}} = \alpha \cdot \mathbf{P} + (1 - \alpha) \cdot \mathbf{I},
\end{equation}
where $\alpha \in [0, 1]$ denotes the transparency level of the visual prompt.

Using the blended image with the highlighted regions and the accompanying detailed text description, we perform autoregressive language modeling to maximize the likelihood of generating the tokens of the ground-truth answer $T_x$:
\begin{equation}
\begin{aligned}
&P(\mathbf{T}_x \mid \hat{\mathbf{I}}, \mathbf{T}_{\text{instruct}}) = \\
&\prod_{i=1}^{L} P_{\theta}(x_i \mid \hat{\mathbf{I}}, \mathbf{T}_{\text{instruct}}, \mathbf{T}_{x, < i})
\end{aligned}
\end{equation}
where $\theta$ represents the trainable parameters, $X_{instruct}$ is the text instruction for generating the target description, $L$ is the sequence length of the answer $T_x$, and $T_{x,<i}$ denotes all the answer tokens before the current prediction token $x_i$, where $i$ denotes the steps during text token generation.
\subsection{Entity Recognition}
Since the Med-VLM requires regions of interest (ROI) to establish region-level cognition and understanding, it is crucial to obtain effective visual prompts for medical images. A key challenge we face is generating accurate region-level annotations for medical images. We adopted a two-step approach: first, extracting region-level entities and then feeding these entities into a visual prompt extractor to generate the visual prompt. In the medical visual question answering task, we start by analyzing the question to identify potential region-level entities that could aid in answering it. To generate these entities, we employ an LLM, which offers greater flexibility and can be easily guided with prompts to produce helpful entities compared to other entity recognition models. For each question $Q$ in the visual question answering task, we prompt the LLM to identify a set of entities: 
\begin{equation}
\mathcal{E} = \text{LLM}(Q, T_{\text{prompt}})
\end{equation}
where $\mathcal{E}$ represents the set of entities the LLM captures from question $Q$, and $T_{\text{prompt}}$ represents the prompt for entity extraction. The extracted entities may include specific organs or diseases mentioned in $Q$ or more general terms for potential relevant organs and diseases. They are the preliminary for generating visual prompts using the visual prompt extractor.

\subsection{Visual Prompt Position Extraction}
In the previous step, we identified region-level entities that could potentially assist in answering the question. The next step involves using a visual prompt extractor to obtain the visual prompt coordinates, utilizing these entities and the corresponding image. Specifically, we employed the Grounding DINO model~\cite{liu2023grounding}, which is a detection model that supports (image, text) input for open-vocabulary detection of entities described by the provided text within the image. Grounding DINO has shown strong performance in open-vocabulary detection on natural images. However, due to the extensive use of specialized terminology for organs and lesions in the medical field, the existing Grounding DINO model struggles to accurately detect region-level entities in the medical images domain. To address this, we first fine-tune Grounding DINO for the medical imaging domain.

Using a medical dataset consisting of images, texts, and coordinates, we adapt Grounding DINO to the specific requirements of medical image detection. Fine-tuning involves applying contrastive loss~\cite{radford2021learning} between predicted objects and language tokens for classification, alongside L1 loss and GIoU loss~\cite{rezatofighi2019generalized} for bounding box regression.

Following fine-tuning, we obtain a Grounding DINO model capable of accurately detecting medical entities. By inputting the region-level entity names into the model, we extract the visual prompt coordinates corresponding to the relevant regions in the medical images:

\begin{equation}
\textbf{P} = \text{G-DINO}(\textbf{I}, \mathcal{E})
\end{equation}

\subsection{Visual Prompt Generation}

We incorporate the visual prompts detected by Grounding DINO into medical images through the alpha blending technique outlined in Equation 1. Since manual annotation of medical images is a common practice in clinical settings, it is essential for our model to be capable of recognizing and interpreting the diverse types of visual prompts frequently employed in these contexts. To address this, we introduce a set of clinically prevalent visual prompts $(\texttt{scribble}, \texttt{rectangle}, \texttt{ellipse})$ during training to enhance the model’s capability in recognizing different annotation types. For each set of coordinates, one of these shapes is randomly selected and applied through alpha blending, resulting in images that contain diverse forms of visual prompt annotations.

By integrating these region-specific visual prompts into the image, we steer the model’s attention toward the relevant areas indicated in question $Q$. To facilitate this, we extract two key attributes $(\texttt{color},\texttt{category})$ of the visual prompt to describe the visual annotation and guide the model’s focus to the appropriate region.

%% file: structure/4_experiment.tex
\section{Experiment}

\begin{table*}[h]
\centering
\caption{Performance of MedVP-LLaVA on three medical visual question answering datasets. Following the routing in LLaVA-Med~\cite{li2024llava}, we report the recall value in column Open for open-set questions. For closed-set questions, we report the accuracy in the Closed column. The best and the second best results are $\textbf{bold}$ and \underline{underlined}. 
}
\vspace{-1em}
\small 
\setlength{\tabcolsep}{10pt} 
\begin{tabular}{p{6cm}|cc|cc|c}
\toprule
\multirow{2}{*}{Models} & \multicolumn{2}{c|}{\textbf{VQA-RAD}} & \multicolumn{2}{c|}{\textbf{SLAKE}} & \textbf{PMC-VQA} \\ 
& Open   & Closed & Open  & Closed & Closed \\ \midrule
\multicolumn{6}{l}{\emph{Zero Shot}} \\ \midrule 
BLIP-2~\cite{li2023blip}                  &  17.5     & 67.7      & 26.9     & 52.4      & 24.3   \\ 
Qwen-VL-Chat~\cite{bai2023qwen} & - & 47.0 & - & 56.0 & 36.6 \\
Open-Flamingo~\cite{awadalla2023openflamingo}           & 15.4      & 61.4      & 13.8     & 29.5      & 26.4   \\ 
LLaVA~\cite{liu2024improved}	&20.7	&59.1	&26.8	&50.2&	29.7 \\
RadFM~\cite{wu2023towards} & - & 50.6 & - &34.6 & 25.9  \\
Med-Flamingos~\cite{moor2023med}           & 20.4      & 71.9      & 16.9     & 49.5      & 27.2   \\ 
LLaVA-Med~\cite{li2024llava}               & 28.2  & 61.4   & 39.2 & 52.2  &  32.9     \\ 

\midrule
\multicolumn{6}{l}{\emph{Finetuned on the Respective Dataset}} \\ \midrule
PMC-CLIP~\cite{lin2023pmc}                & 52.0     & 75.4   & 72.7  & 80.0     & 37.1 \\
MedVInT-TE~\cite{zhang2023pmc}               & 69.3   & 84.2   & \underline{88.2}  & 87.7   & 39.2   \\ 
MedVInT-TD~\cite{zhang2023pmc}              & 73.7   & \underline{86.8}   & 84.5  & 86.3   & 40.3   \\ 
LLaVA-Med~\cite{li2024llava}    & 72.2      & 84.2      & 70.9    & 86.8      & 42.8   \\ 
HuatuoGPT-Vision~\cite{chen2024huatuogpt}  & - & 68.9 & - & 84.1 & 57.3 \\
LLaVA-Med++~\cite{xie2024medtrinity}             & \underline{77.1}   & 86.0     & 86.2  & \underline{89.3}   & \textbf{61.9}      \\

MedVP-LLaVA (ours)      & \textbf{89.3}  & \textbf{97.3}  & \textbf{91.6} & \textbf{92.9}  & \underline{58.3}  \\ \bottomrule
\end{tabular}
\label{tab:model_comparison}
\end{table*}

\subsection{Experiment Setup}

\textbf{Base Model.} 
We utilize ViP-LLaVA 7B~\cite{cai2024vip} as our base model. ViP-LLaVA~\cite{cai2024vip} was initially pre-trained on natural image datasets and allows for user interaction by incorporating visual prompts into the images, enhancing the model’s ability to interpret and respond to inputs. We use it as our base model because it has a better capability of identifying regions in marked regions 

\noindent\textbf{Knowledge Pre-training of Med-VLM.} 
To inject medical knowledge into our model, we pre-train ViP-LLaVA 7B using a subset of the MedTrinity-20M dataset~\cite{xie2024medtrinity}, which consists of triplets in the format {image, ROI, description}. Each ROI (region of interest) corresponds to an abnormality annotated with a bounding box. The descriptions provide a multi-granular textual explanation, covering disease or lesion type, imaging modality, region-specific details, and inter-regional relationships. This pre-training on medical images equips ViP-LLaVA with domain-specific knowledge, enabling the model to focus on region-specific information in medical contexts.

\noindent\textbf{Finetuning of Visual Prompts Extractor.} 
We employ Grounding DINO\cite{liu2023grounding} as our visual prompt extractor, which processes text inputs to identify corresponding regions in medical images. To adapt Grounding DINO for annotation tasks in the medical domain, we fine-tune it using a combination of the SLAKE training dataset and a subset of the SA-Med2D dataset\cite{ye2023sa}. The SLAKE~\cite{liu2021slake} dataset has region-level annotations of different organs and diseases on the radiological image. The SA-Med2D dataset~\cite{ye2023sa} includes over 20,000 images from various modalities, such as MRI, CT, ultrasound, PET, X-ray, fundus, and endoscopy. The annotated masks are converted into bounding box coordinates, covering a wide range of organs and disease types. 

\noindent\textbf{Choice of Visual Prompts.} 
We utilize three types of visual prompts for image annotation: scribble, rectangle, and ellipse, chosen for their frequent application in medical imaging. During fine-tuning, different visual prompt shapes are randomly applied to the images, allowing the model to learn how to interpret and respond to a variety of prompt shapes, thereby improving its robustness and flexibility. The Figure~\ref{fig:visualize_visual_prompts} in appendix shows examples of different types of visual prompts.

\subsection{Results}
In this section, we assess the performance of MedVP-LLaVA across three distinct medical VQA datasets, comparing our approach against several state-of-the-art methods in the field.
\subsubsection{Evaluation of the Benchmarks }
We evaluate our method on three medical VQA datasets: SLAKE~\cite{liu2021slake}, VQA-RAD~\cite{lau2018dataset}, and PMC-VQA~\cite{zhang2023pmc}. Both SLAKE and VQA-RAD contain two types of questions: Open and Closed. PMC-VQA is a multi-modal, multiple-choice VQA dataset. 
As shown in Table 1, our approach either outperforms or achieves comparable results compared with the current state-of-the-art (SOTA) methods. On the SLAKE and VQA-RAD datasets, our method achieves significant improvements. Specifically, on the VQA-RAD dataset, our method demonstrates an improvement of over 10\% on both the open and closed metrics. On the SLAKE dataset, our approach also achieves a significant average improvement of approximately 4\% compared to the best-performing models. The fine-tuned MedVP-LLaVA shows particularly strong gains on the closed-set questions in the VQA-RAD dataset. 
For the PMC-VQA dataset, where each question requires selecting the correct answer from four options, our method also achieves the second-best performance, surpassing a range of advanced models in the medical VQA setting, except LLaVA-Med++~\citep{xie2024medtrinity}, which benefits from training on a larger synthetic dataset



\subsection{Analysis}
In this section, we analyze the impact of visual prompts on the Medical VQA task and evaluate the performance of MedVP-LLaVA under different visual prompt shape configurations. Additionally, we visualize the model’s region-level attention maps to illustrate how it attends to specific Regions of Interest (ROIs) during inference.

\subsubsection{Effectiveness of Visual Prompts in the Fine-tuning Stage}
During the fine-tuning phase, visual prompts are integrated into medical images, and the corresponding enhanced text inputs are utilized to fine-tune the medical vision-language model. To quantify the performance improvement attributed to visual prompts, we conduct comparative experiments with and without their inclusion during fine-tuning. Our findings indicate that visual prompts play a pivotal role in enhancing performance on Medical VQA tasks.

\begin{table}[t]
\centering
\scriptsize
\caption{Ablation study on the impact of using visual prompts during fine-tuning on downstream medical VQA datasets. ‘VP' in the first column refers to visual prompts. Both the open and close set accuracy are reported.}
\begin{tabular}{p{1.6cm}|cc|cc|c}
\toprule
\multirow{2}{*}{MedVP-LLaVA} & \multicolumn{2}{c|}{\textbf{VQA-RAD}} & \multicolumn{2}{c|}{\textbf{SLAKE}} & \textbf{PMC-VQA} \\ 
                  & Open     & Closed   & Open   & Closed   & Closed    \\ \midrule
w/o VP   & 79.81     & 92.92     & 72.56   & 87.23     & 54.14     \\ 
w VP       & \textbf{80.41}    & \textbf{97.27}    & \textbf{79.22}      & \textbf{92.88}    & \textbf{58.30}      \\ 
\bottomrule
\end{tabular}
\label{tab:enhancement_after_vp}
\end{table}

As shown in Table~\ref{tab:enhancement_after_vp}, incorporating visual prompts during fine-tuning leads to a substantial improvement in test performance compared to models trained without them. These results underscore the importance of Med-VLM’s capability to capture and analyze local visual information in the context of Medical VQA.

\begin{figure*}[h]
  \centering
  \caption{Visualization results of our cross-attention map. The red boxes are the visual prompts generated by our grounding model. As illustrated, the yellow area indicates high attention values and largely overlaps with the visually prompted area.}
  \includegraphics[width=\textwidth]{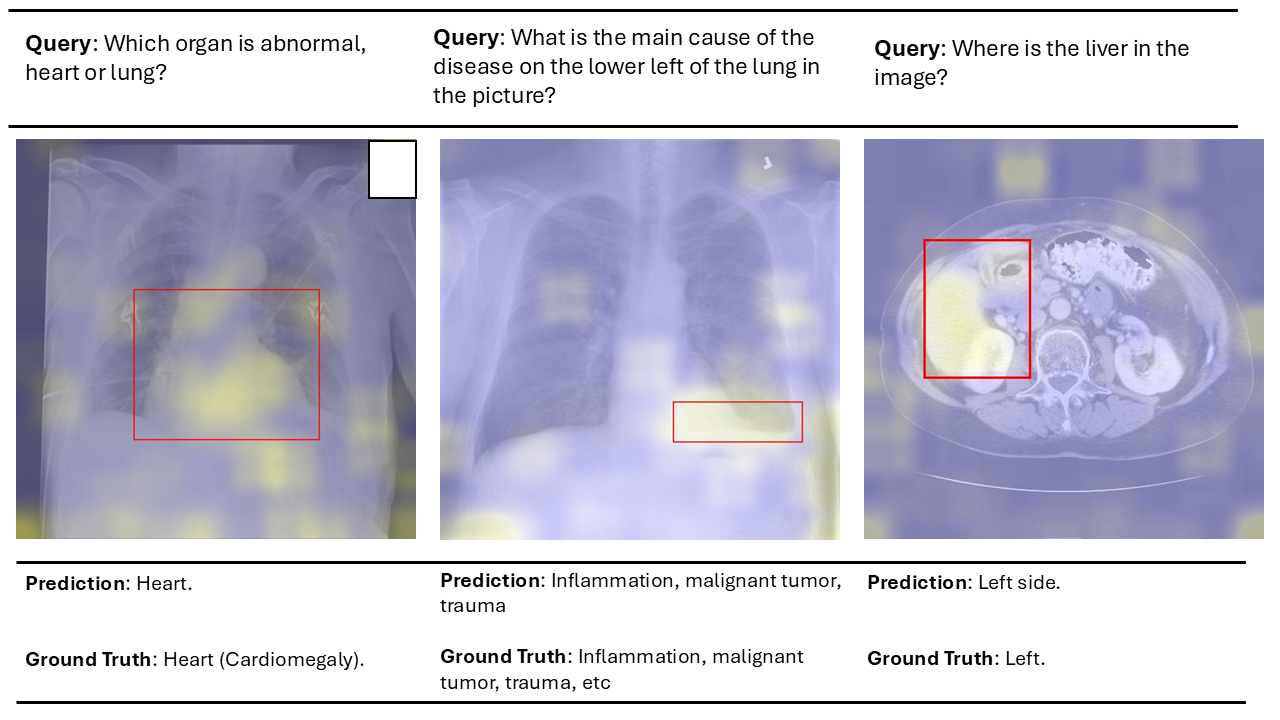}

  \label{fig:visualize}
  \vspace{-1em}
\end{figure*}

\begin{table}[htbp]
\centering
\caption{Analysis of robutsness of the model with limited visual prompts in the testing phase. The ratio column denotes the ratio of visual prompts being kept in the testing phase inference stage. We report both the accuracy of the open-set and closed-set questions.}
\resizebox{\linewidth}{!}{\begin{tabular}{lcccccc}
\toprule
\multirow{2}{*}{Ratio} & \multicolumn{2}{c}{\textbf{VQA-RAD}} & \multicolumn{2}{c}{\textbf{SLAKE}} & \textbf{PMC-VQA} \\
& Open & Closed & Open & Closed & Closed \\
\midrule
20\% &  78.86 & 96.88 & 78.22 & 89.45 &57.94 \\
40\% &  78.35 & 96.88 & 78.37 & 90.35 &58.01 \\
60\% &  78.86 & 97.27 &78.69 & 89.84 & 58.13 \\
80\% & 78.35 & 96.10 &78.69 & 91.73 &  58.10 \\
100\% & \textbf{80.41} & \textbf{97.27} & \textbf{79.22} & \textbf{92.88}  & \textbf{58.30}     \\
\bottomrule
\end{tabular}
}
\label{tab:labeling_ratios}
\end{table}
\subsubsection{Robustness of the Model during Inference with Limited Visual Prompts}

During the fine-tuning phase, our framework leverages visual prompts to guide the model toward specific regions, enabling it to focus more effectively on relevant areas when performing Medical VQA tasks. To evaluate whether the model has learned to naturally and robustly attend to relevant regions with guided visual prompts, we design an experiment to examine the impact of progressively reducing the proportion of visual prompts generated by Grounding DINO during inference.

For each dataset’s test set, where Grounding DINO has provided visual prompt coordinates, we systematically remove a certain percentage of these prompts at random, retaining only 80\%, 60\%, 40\%, and 20\% of the original visual prompts. It is important to note that the models used in this experiment were fully fine-tuned with visual prompts during training, and the random removal of visual prompts during inference is entirely independent of the prompts utilized in the fine-tuning stage.


As shown in Table~\ref{tab:labeling_ratios}, reducing the proportion of visual prompts during inference leads to a slight degradation in performance compared to the fully annotated setting. However, the overall performance remains relatively stable, demonstrating the robustness of the model. We attribute this robutsness to the effectiveness of visual prompts during fine-tuning, which facilitate the internalization of spatial attention patterns. Given the inherent heterogeneity of certain medical VQA datasets, the distribution of Regions of Interest (ROIs) across training and test sets often exhibits substantial similarity. Consequently, models fine-tuned with visual prompts as explicit attention guidance inherently attend to similar ROIs, even in the absence of explicit visual prompts during inference. This suggests that visual prompts not only direct the model’s attention during training but also instill a learned focus on relevant regions, enabling it to prioritize key visual details effectively. As a result, even with a reduced number of visual prompts during inference, the model maintains its capacity to generate accurate responses by leveraging its learned spatial attention mechanisms.


\begin{table}[htbp]
\centering
\footnotesize
\caption{Performance of different visual prompt types on three medical VQA datasets during inference. The ‘Mix’ row represents the use of a combination of the three visual prompt types. We report both the accuracy of the open-set and closed-set questions.}
\setlength{\tabcolsep}{4pt} 
\renewcommand{\arraystretch}{1.2} 
\begin{tabular}{l|cc|cc|c}
\toprule
\multirow{2}{*}{Category}   & \multicolumn{2}{c|}{\textbf{VQA-RAD}} & \multicolumn{2}{c|}{\textbf{SLAKE}} & \multicolumn{1}{c}{\textbf{PMC-VQA}} \\ 
   & Open       & Closed        & Open       & Closed        & Closed       \\ \midrule
Scribble   & 79.38           & 96.10          & 80.62         & 91.82        & 58.19        \\ 
Rectangle  & 81.90           & 96.40          & 78.29         & 91.35        & 58.24        \\ 
Ellipse    & \textbf{82.47}  & 96.80          & 79.06         & 92.54        & 58.12        \\ \midrule
Mix        & 80.41           & \textbf{97.27} & \textbf{79.22} & \textbf{92.88} & \textbf{58.30} \\ \bottomrule
\end{tabular}
\label{tab:vp_shapes}
\end{table}

\subsubsection{Performance Comparison on Different Types of Visual Prompts} 
We evaluated the performance of our model on Medical VQA across different types of visual prompts. During the inference stage, as shown in Table~\ref{tab:vp_shapes}, we constrain the model to utilize only a single type of visual prompt for the Medical VQA task. The results indicate that ellipse-shaped visual prompts achieve the highest performance, followed by rectangular prompts, with scribble prompts yielding the lowest accuracy. This performance variation can be attributed to differences in visual saliency and spatial coverage—both rectangular and elliptical prompts tend to highlight a broader area and exhibit greater visual prominence compared to scribble-based prompts. To enhance the model’s robustness against variations in prompt shape, our framework incorporates a combination of three visual prompt types during both fine-tuning and inference. This multi-prompt strategy enables the model to generalize more effectively across different visual prompt configurations. As a result, even when evaluated using a single type of visual prompt, the model maintains relatively stable performance, demonstrating its adaptability to diverse forms of visual guidance.

\subsubsection{Visualization of Model's Region-level Attention}
To further investigate whether the model has correctly established region-level attention and understanding, we visualized MedVP-LLaVA’s attention on the medical images, as shown in Figure~\ref{fig:visualize}. Following the setting in this work~\cite{yu2024attentionpromptingimagelarge}, we visualize the cross-attention map between the generated tokens and the visual embedding tokens. The regions colored yellow to show the regions that affect more on the model's response. In the first example, the heatmap shows that our model accurately focuses on the region of interest. Additionally, in the rightmost figure, we can see that the model correctly attends to the liver.
This attention map illustrates the regions of the image that the model focuses on while generating responses to the queries. As illustrated in the figure, we can conclude that after our visual-prompt-guiding fine-tuning, our model can constrain its attention to our visual prompts area. 

\section{Impact}
Our approach integrates visual prompts to facilitate the Medical VQA task, not only enhancing model performance but also offering clinical relevance for medical image analysis. During training, visual prompts serve as attention guidance, directing the model to focus on the most clinically relevant regions, thereby improving both the efficiency and effectiveness of the learning process. In the inference stage, visual prompts provide contextual reference points, enhancing the interpretability of the model’s responses while contributing to improved safety and reliability in medical decision-making. Furthermore, the proposed framework is easily adaptable to real-world clinical scenarios, where medical professionals can directly annotate visual prompts onto medical images, seamlessly integrating them into the workflow. These factors collectively underscore the potential of our approach in advancing medical imaging applications, improving both automated analysis and human-AI collaboration.

%% file: structure/5_limitation.tex
\section{Limitations}
During the knowledge pre-training stage, the dataset we utilize contains AI-generated visual markers to guide the model's attention. These contents might not be factually accurate, due to the limitation of grounding models we adopted. Therefore, we perform a small-scale error analysis (\ref{error_analysis}) and a human evaluation (\ref{quality_eval}) on the AI-generated visual prompts, with detailed results provided in the appendix. Due to the high cost and scarcity of human-annotated medical image data, we rely on AI-generated annotations to augment our training corpus. In future work, we aim to develop more rigorous data filtering and validation strategies and integrate retrieval-augmented generation (RAG) techniques into our model to mitigate factual inaccuracies.

%% file: structure/6_conclusion.tex
\section{Conclusion}
In this work, we propose the MedVP method, which introduces explicit visual prompts into medical images to guide Med-VLMs in focusing on specific Regions of Interest (ROIs) for medical VQA tasks. Our approach includes a medical visual prompt extraction process. First, we use LLM to identify medical entities from the questions. Next, we train a grounding model to locate the ROIs in the images based on the extracted entity names. Finally, we blend the visual prompts with the medical images. Our MedVP-LLaVA achieves significant performance improvements across three Medical VQA datasets, demonstrating the effectiveness of incorporating visual prompts to enhance medical VLMs’ capability in medical question answering.

%% file: structure/appendix.tex
\appendix

\section{Data}
\label{sec:appendix_a}

\subsection{Data Statistics}

The quantities of all the data sets are shown in Table~\ref{tab:multi_datasets}. During the knowledge pre-training stage, we extract a subset of Med-Trinity datasets to fine-tune our model. This subset includes images from SLAKE, VQA-RAD, and PATH-VQA.

During fine-tuning, we only use the training set data from the SLAKE, VQA-RAD, and PMC-VQA datasets.

Table~\ref{tab:dino_image_statistics} shows the number of images from different modalities we selected from the SA-Med-2D dataset when fine-tuning the Grounding DINO.

\subsection{Involved Datasets}
We leveraged three publicly available medical VQA datasets: VQA-RAD, SLAKE, and PMC-VQA.

\begin{itemize}
    \item \textbf{SLAKE}: SLAKE is a bilingual dataset in English and Chinese, comprising 642 images and 14,028 question-answer pairs, designed for training and evaluating Med-VQA systems.
    \item \textbf{VQA-RAD}: A radiology-focused VQA dataset that includes radiological images and corresponding questions aimed at assessing the model's performance in answering domain-specific questions about various types of radiological findings.
    \item \textbf{PMC-VQA}: A large-scale dataset constructed from PubMed Central (PMC) articles, focusing on medical visual question answering with figures and charts extracted from research papers, testing models on understanding complex visual and textual medical data.
\end{itemize}

\begin{table*}[t]
\centering
\caption{Data statistics for various datasets in different training stages} 
\resizebox{0.9\textwidth}{!}{\begin{tabular}{l|cc|cc|cc|cc}
\toprule
\multirow{2}{*}{Dataset}   & \multicolumn{2}{c|}{\textbf{SLAKE}} & \multicolumn{2}{c|}{\textbf{VQA-RAD}} & \multicolumn{2}{c|}{\textbf{PMC-VQA}} & \multicolumn{2}{c}{\textbf{Path-VQA}}\\ 
   & QA-pair       & Image        & QA-pair   & Image   & QA-pair   &Image  & QA-pair & Image    \\ \midrule
Pre-train (Med-Trinity) & 642  & 642  & 1754    & 1754        & \ding{55} & \ding{55}   & 13371 & 13371       \\ 
Fine-tuning  & 9834   &642     & 1793     & 313            & 227000    & 149000 & \ding{55}     \\ 
Test    & 2094  & 642    & 451     & 203   & 164360        &33430      & \ding{55} & \ding{55}    \\ \midrule

\end{tabular}
}
\label{tab:multi_datasets}
\end{table*}

\begin{table*}[t]
\centering
\caption{Image count statistics across different imaging modalities when fine-tuning Grounding DINO} 
\resizebox{0.9\textwidth}{!}{\begin{tabular}{l|cccccccc}
\toprule
\multirow{1}{*}{Count} & \multicolumn{1}{c}{\textbf{MR}} & \multicolumn{1}{c}{\textbf{CT}} & \multicolumn{1}{c}{\textbf{Xray}} & \multicolumn{1}{c}{\textbf{Endoscopy}} & \multicolumn{1}{c}{\textbf{Dermoscopy}} & \multicolumn{1}{c}{\textbf{Fundus}} & \multicolumn{1}{c}{\textbf{Ultrasound}} & \multicolumn{1}{c}{\textbf{PET}} \\ 
\midrule
Image Count & 4000 & 4000 & 3192 & 1863 & 866 & 1470 & 1645 & 1000 \\ 
\bottomrule
\end{tabular}
}
\label{tab:dino_image_statistics}
\end{table*}

\section{Visualization of Visual Prompts}
\label{sec:appendix_b}
\begin{figure*}[htbp]
  \centering
  \caption{Visualization of different types of visual prompts integrated into medical images. The images depict various shapes, including scribbles, rectangles, and ellipses, with varying line thicknesses.}
  \includegraphics[width=\textwidth]{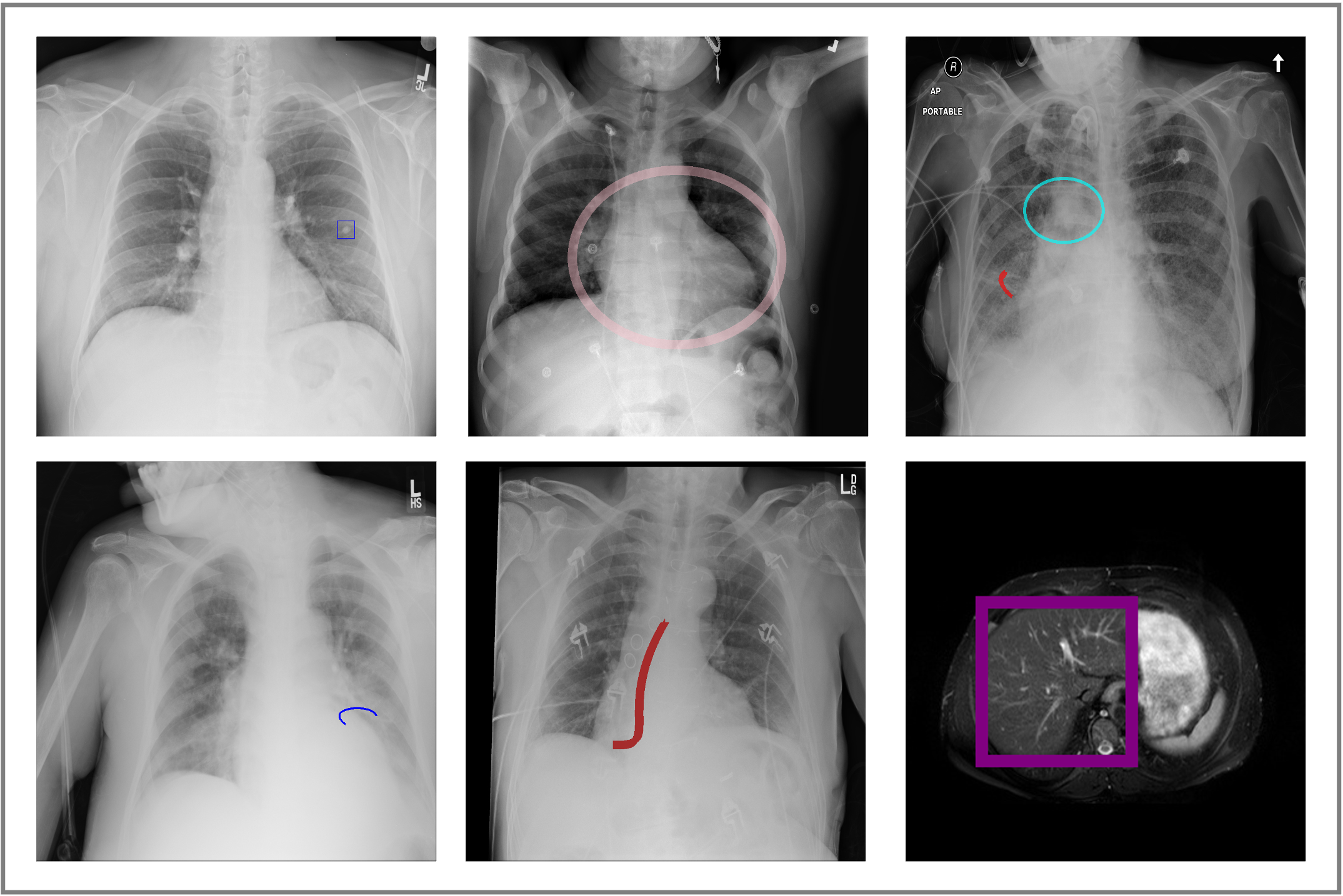}

  \label{fig:visualize_visual_prompts}
  \vspace{-1em}
\end{figure*}

In Figure~\ref{fig:visualize_visual_prompts}, we present the visualization results of the integrated visual prompts in the images, demonstrating three types of visual prompts with varying sizes and levels of transparency.

\section{Implementation Details}
\label{sec:appendix_c}
All experiments were conducted on RTX 4090 and H100 GPUs. For region-level medical domain alignment, we used a learning rate of 2e-5 and a batch size of 64. When detecting bounding boxes with Grounding DINO, we set the prediction score threshold to 0.2 and the batch size to 24.

For downstream fine-tuning, we applied a learning rate of 2e-5, a batch size of 128, and a warm-up ratio of 0.03. The fine-tuning process on 4 RTX 4090 GPUs took approximately 6 hours for SLAKE, 2 hours for VQA-RAD, and 20 hours for PMC-VQA. Grounding DINO fine-tuning, also on 4 RTX 4090 GPUs, took around 9 hours.
\section{Instructions and Prompts}
\begin{figure}[htbp]
  \centering
  \caption{Instructions for inference with the integrated visual prompts.}
  \includegraphics[width=\columnwidth]{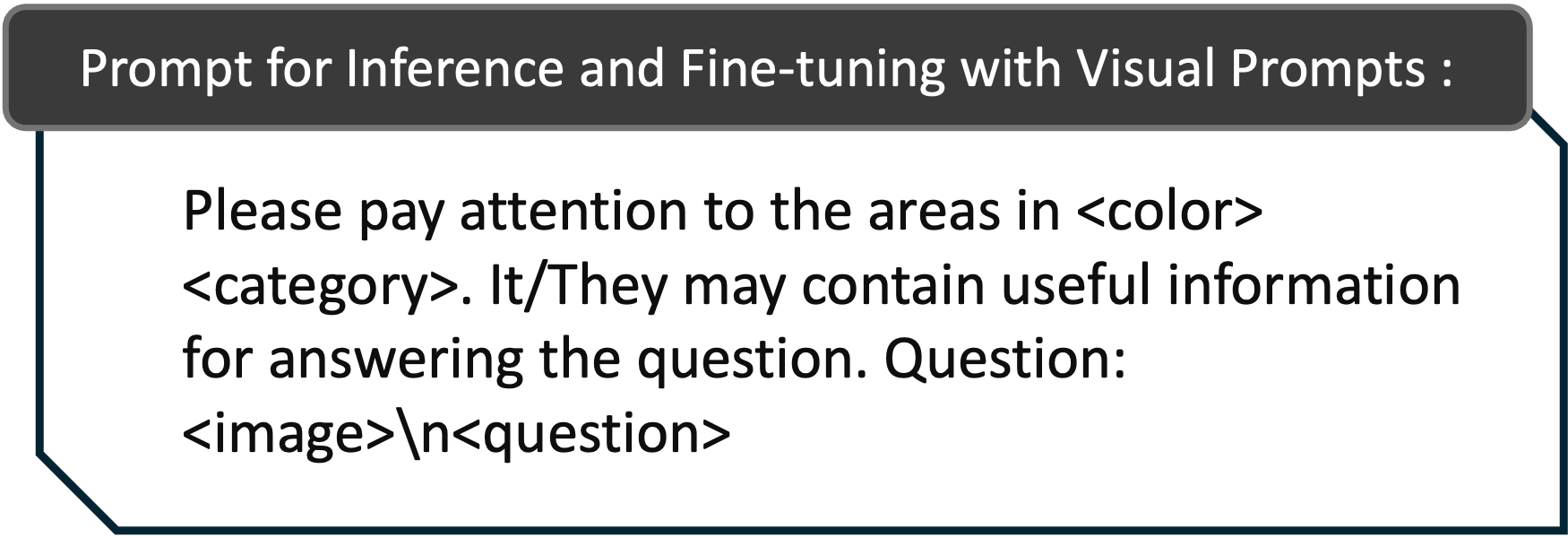}

  \label{fig:prompt1}
  \vspace{-1em}
\end{figure}

\begin{figure}[htbp]
  \centering
  \caption{Instructions for generating entities in the entity recognition task.}
  \includegraphics[width=\columnwidth]{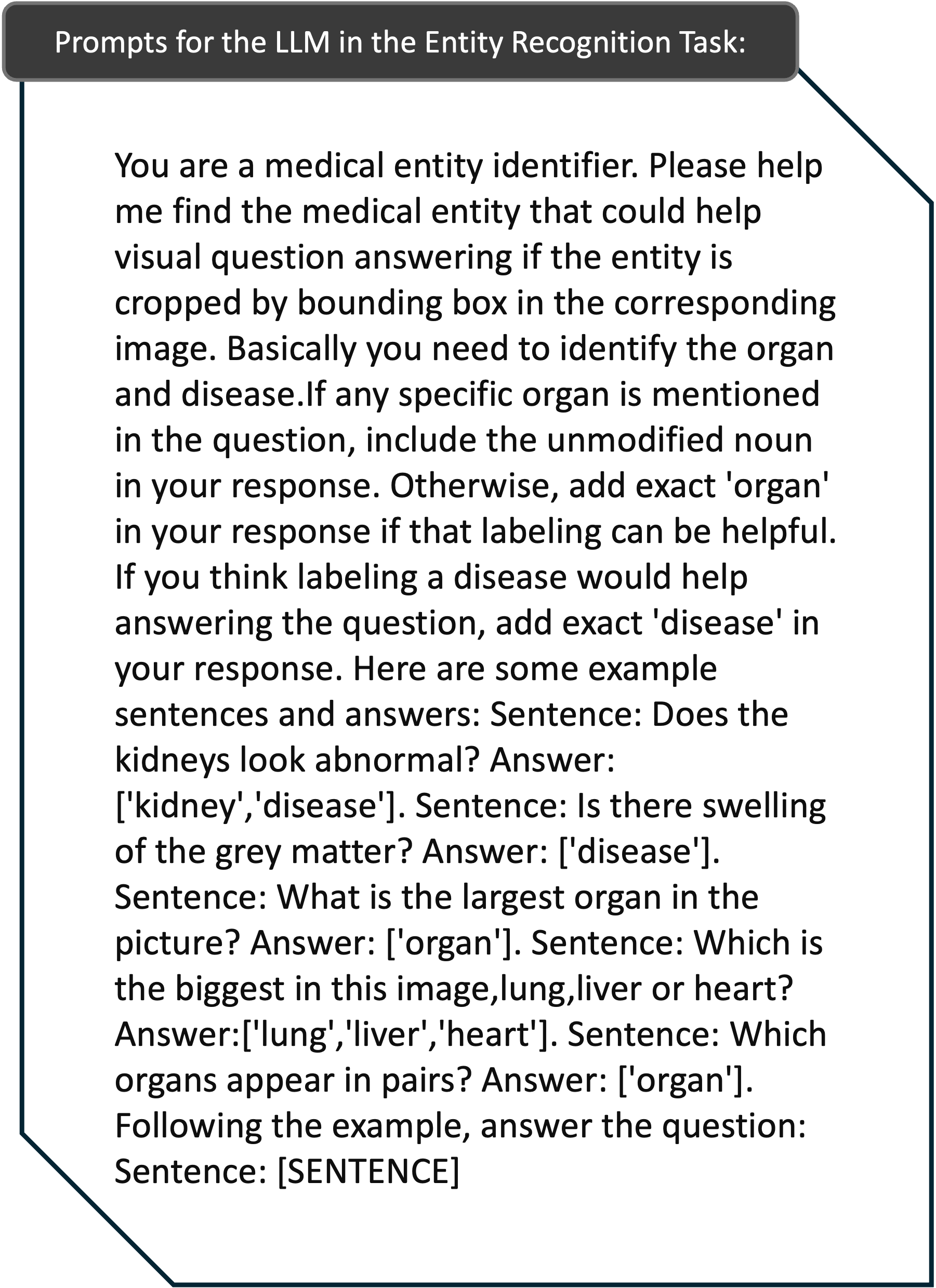}

  \label{fig:prompt2}
  \vspace{-1em}
\end{figure}

We show the prompt used for the entity recognition task and the instruction to Med-VLM when performing downstream fine-tuning and inference in Figure~\ref{fig:prompt1} and Figure~\ref{fig:prompt2}.
\label{sec:appendix_d}


\section{Error Analysis}
\label{error_analysis}

\begin{figure*}[htbp]
  \centering
  \caption{Instances where visual prompts may mislead the medical vision-language model.}
  \includegraphics[width=\textwidth]{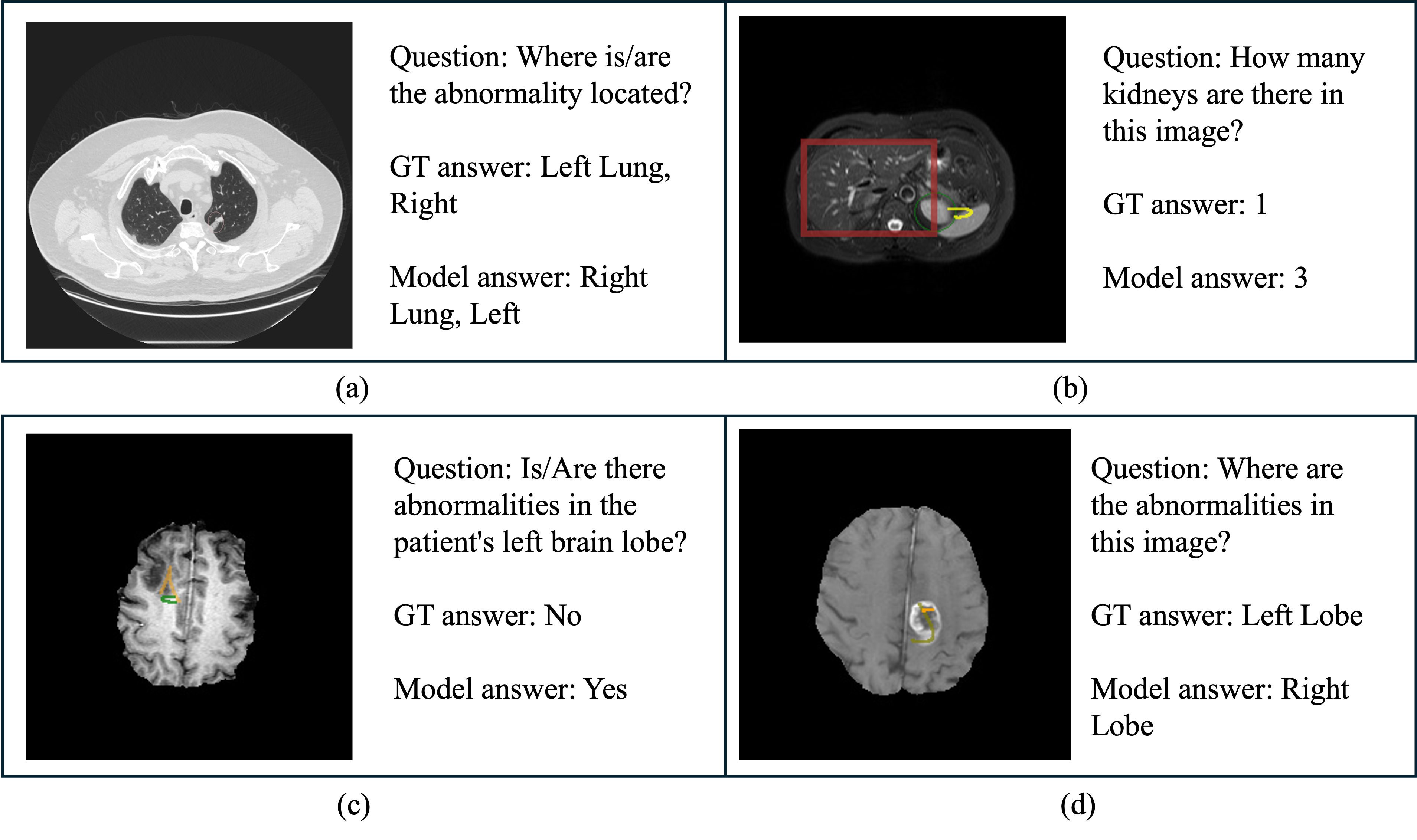}

  \label{fig:error_analysis}
  \vspace{-1em}
\end{figure*}

While visual prompts can enhance the model’s accuracy in question answering, we have also identified instances where they inadvertently lead to incorrect judgments. To better understand these failure cases, we conducted a systematic analysis and categorized the underlying causes of errors into the following types:

First, we found that some visual prompts were too small or used colors that were not easily perceived by the model, leading to errors. For example, in Figure~\ref{fig:error_analysis} (a), the visual prompt is extremely small with very fine lines, making it less noticeable to the model and potentially leading to error.

Secondly, for certain counting questions, the number of visual prompts provided can occasionally be misleading. For instance, as illustrated in Figure~\ref{fig:error_analysis} (b), the model may predict the number of kidneys based solely on the number of visual prompts, regardless of the actual content of the image.

Thirdly, visual prompts can mislead the model into providing incorrect answers to questions regarding the existence of abnormalities. For instance, in Figure~\ref{fig:error_analysis} (c), even when no abnormality is present in the image, the presence of a visual prompt may cause the model to incorrectly answer "Yes."

Finally, the model can sometimes rely on the location of visual prompts when answering location-related questions. In Figure~\ref{fig:error_analysis} (d), if the ground truth specifies the left lobe but the visual prompt is mistakenly placed on the right lobe, the model may erroneously predict the right lobe as the answer.

\label{sec:appendix_e}

\section{Human Evaluation}
\label{quality_eval}
\begin{figure*}[htbp]
  \centering
  \caption{Visualization of Grounding DiNO predicting results on different datasets.}
  \includegraphics[width=\textwidth]{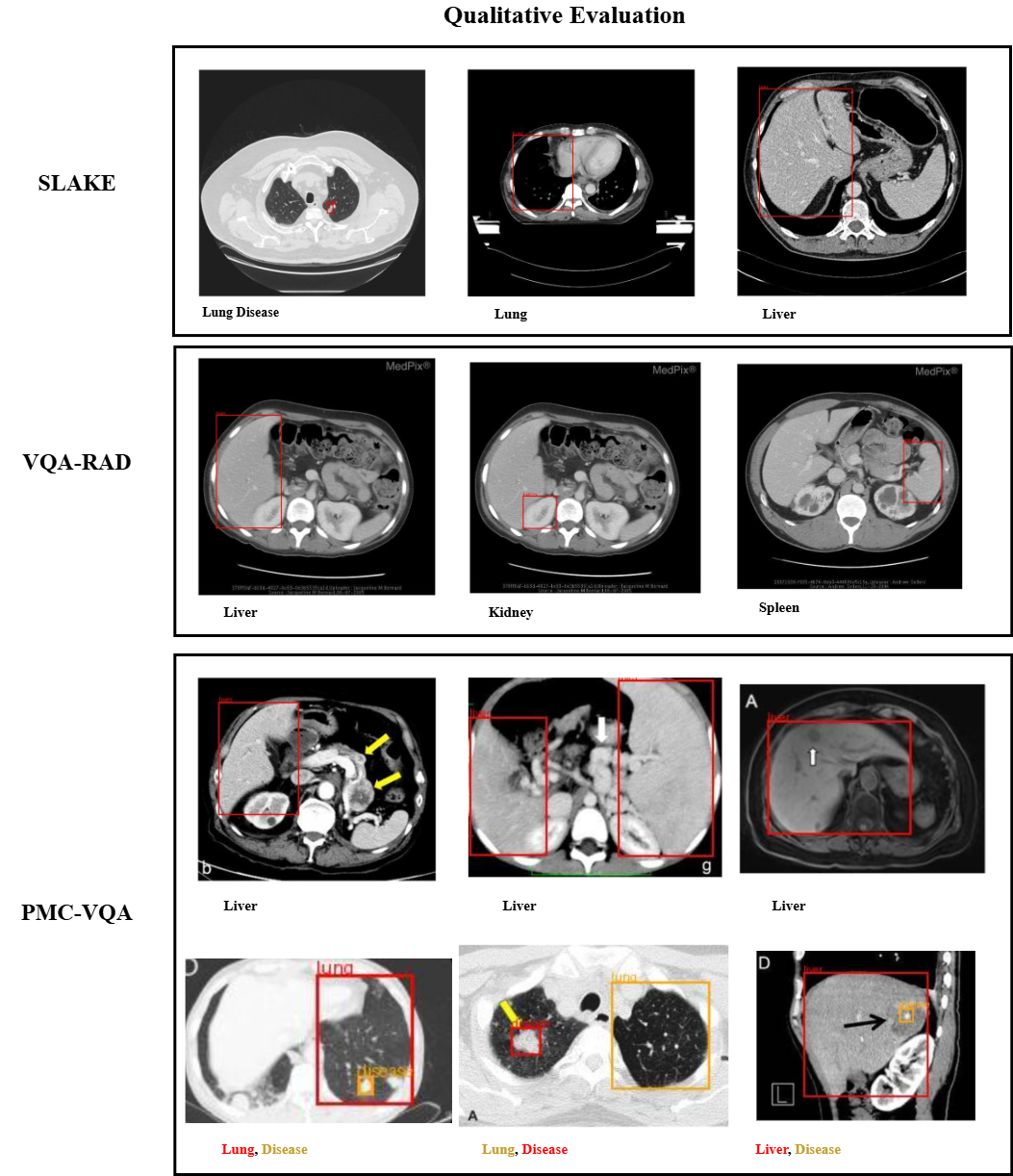}

  \label{fig:qualitative_good}
  \vspace{-1em}
\end{figure*}

As illustrated in Figure~\ref{fig:qualitative_good}, we present the grounding results of our fine-tuned grounding model. We fine-tune the Grounding DINO model on the training sets of the SLAKE~\citep{liu2021slake} and Sa-Med2D-20k~\citep{ye2023sa} datasets. By providing the name of the target organ or other medical entities as a prompt, the Grounding DINO model automatically generates bounding boxes for the corresponding regions in the image.
As shown in the figure, the model performs well in most cases, accurately recognizing common organs and abnormal regions as needed.

\label{sec:appendix_f}

